%% file: main.tex
\documentclass{article}
\usepackage[preprint]{colm2026_conference}

\usepackage{microtype}
\usepackage{hyperref}
\usepackage{url}
\usepackage{booktabs}
\usepackage{amsmath,amssymb}
\usepackage{graphicx}
\usepackage{xcolor}
\usepackage{enumitem}
\usepackage{subcaption}
\newcommand{\nicefrac}[2]{#1/#2}
\usepackage{caption}
\usepackage{pifont}
\usepackage{lineno}
\usepackage[most]{tcolorbox}
\usepackage{colortbl}

\definecolor{darkblue}{rgb}{0, 0, 0.5}
\definecolor{hmcolor}{rgb}{0.8, 0.2, 0.0}

\hypersetup{colorlinks=true, citecolor=darkblue, linkcolor=darkblue, urlcolor=darkblue}

\title{
SenseMath: Do LLMs Have Number Sense? \\
Evaluating Shortcut Use, Judgment, and Generation}

\author{
  \begin{tabular}[t]{c}
  \bf Haomin Zhuang, Xiangqi Wang, Yili Shen, Ying Cheng, Xiangliang Zhang \\
  University of Notre Dame \\
  \texttt{\{hzhuang2, xzhang33\}@nd.edu}
  \end{tabular}
}

\begin{document}

\ifcolmsubmission
\linenumbers
\fi

\maketitle

\begin{abstract}
\input{sections/0_abstract}
\end{abstract}

\input{sections/1_introduction}

\input{sections/2_related_work}
\input{sections/3_benchmark}
\input{sections/4_experiments}
\input{sections/5_conclusion}

\section*{Acknowledgements}
This work is supported by the NSF award \#2321054.

\bibliography{references}
\bibliographystyle{colm2026_conference}

\appendix

\section{Example Items}
\label{sec:appendix_examples}

Table~\ref{tab:examples} shows example \textsc{SenseMath} items at $d{=}2$ for four categories.

\input{figures/TABLE_examples.tex}

\section{Prompt Templates}
\label{sec:appendix_prompts}

\paragraph{Use-level prompts.} Three prompting conditions for solving \textsc{SenseMath} items:

\begin{tcolorbox}[width=0.9\linewidth, colback=blue!3, colframe=blue!40, arc=2pt, title={\textbf{CoT prompt}}]
\small Please solve the following multiple-choice problem. Show your step-by-step reasoning, then provide your final answer as a single capital letter (A, B, C, or D) inside \textbackslash boxed\{\}.
\end{tcolorbox}

\begin{tcolorbox}[width=0.9\linewidth, colback=orange!3, colframe=orange!40, arc=2pt, title={\textbf{NS prompt}}]
\small Please solve the following multiple-choice problem using only easy calculations. Rely on your mathematical intuition, number sense, and estimation. Show brief reasoning, then provide your final answer as a single capital letter (A, B, C, or D) inside \textbackslash boxed\{\}.
\end{tcolorbox}

\begin{tcolorbox}[width=0.9\linewidth, colback=red!3, colframe=red!40, arc=2pt, title={\textbf{Strict prompt}}]
\small Please solve the following multiple-choice problem by performing precise, step-by-step arithmetic. Do NOT use estimation, rounding, or shortcuts. Show every computation digit by digit, then provide your final answer as a single capital letter (A, B, C, or D) inside \textbackslash boxed\{\}.
\end{tcolorbox}

\paragraph{Judge-level prompts.}\leavevmode\par

\begin{tcolorbox}[width=0.9\linewidth, colback=green!3, colframe=green!40, arc=2pt, title={\textbf{J1: Shortcut appropriateness}}]
\small Consider this math problem. Could you solve it significantly faster using mental math or a clever observation, compared to computing it step by step? Answer YES or NO, then explain in one sentence why.
\end{tcolorbox}

\begin{tcolorbox}[width=0.9\linewidth, colback=green!3, colframe=green!40, arc=2pt, title={\textbf{J2: Strategy identification}}]
\small Classify the solution strategy as SHORTCUT or COMPUTATION. SHORTCUT: uses estimation, approximation, pattern recognition, structural insight, cancellation, or benchmark comparison. COMPUTATION: performs standard step-by-step arithmetic.
\end{tcolorbox}

\paragraph{Generate-level prompt.}\leavevmode\par

\begin{tcolorbox}[width=0.9\linewidth, colback=yellow!3, colframe=yellow!50!black, arc=2pt, title={\textbf{G: Problem generation}}]
\small You are given a category of number-sense problems and one example. Your task is to generate a NEW problem pair: one ``strong\_shortcut'' version where a specific mental math shortcut works, and one ``control'' version where the same shortcut does NOT work.

\medskip
Category: \{category\_name\}\\
Description: \{category\_description\}

\medskip
Example:\\
\quad Strong shortcut: \{example\_strong\_question\}\\
\quad\quad $\to$ The shortcut works because: \{explanation\}\\
\quad Control: \{example\_control\_question\}\\
\quad\quad $\to$ The shortcut does NOT work because: \{explanation\}

\medskip
Now generate a NEW problem pair (different numbers). Output in JSON format with fields: question, math\_expression, answer, why\_shortcut\_works/fails.
\end{tcolorbox}

Six deterministic code checks verify generated items: (1) strong answer correctness, (2) control answer correctness, (3) shortcut existence in strong, (4) shortcut absence in control, (5) variant matching, (6) novelty and digit scale.

\paragraph{SU rate judge prompt.}
\label{sec:appendix_su_prompt}\leavevmode\par

\begin{tcolorbox}[width=0.9\linewidth, colback=purple!3, colframe=purple!40, arc=2pt, title={\textbf{Shortcut usage classification (GPT-4.1-mini)}}]
\small Please judge whether the following answer to a math problem uses a shortcut method (simple heuristic approach) rather than relying on precise calculations.

\medskip
\textbf{Shortcut characteristics:} (1) Uses pattern recognition, memorization, or heuristic methods; (2) Skips complete calculation steps; (3) Relies on number properties (parity, number of digits, magnitude estimation) rather than actual computation; (4) Uses approximation or estimation instead of exact values.

\medskip
\textbf{NOT a shortcut:} Answers that rely on precise calculations, show detailed step-by-step computational work, or calculate exact values.

\medskip
Question: \{question\}\\
Options: \{options\}\\
Answer: \{answer\}

\medskip
Respond in JSON: \{``uses\_shortcut'': true/false, ``confidence'': 0--1, ``explanation'': ``...''\}
\end{tcolorbox}

\paragraph{MATH-500 classification prompt.}
\label{sec:appendix_math500_prompt}\leavevmode\par

\begin{tcolorbox}[width=0.9\linewidth, colback=gray!5, colframe=gray!50, arc=2pt, title={\textbf{NS-amenable classification (GPT-4.1-mini)}}]
\small You are classifying whether a math problem can benefit from Number Sense (NS) strategies---intuition, estimation, pattern recognition, magnitude reasoning, or structural shortcuts---or whether it fundamentally requires precise step-by-step algebraic/arithmetic computation.

\medskip
\textbf{NS-AMENABLE:} estimation/approximation, magnitude reasoning, pattern recognition, benchmark comparison, cancellation that avoids full computation.

\medskip
\textbf{COMPUTATION-REQUIRED:} precise multi-step algebra/calculus, exact symbolic manipulation, geometric constructions, problems with no intuitive shortcut.

\medskip
Problem: \{problem\}

\medskip
Reply with exactly one word: NS or COMP
\end{tcolorbox}

\section{Task Specifications}
\label{sec:appendix_tasks}

\paragraph{Heuristic validation.} A deterministic shortcut solver applies category-specific heuristics without exact arithmetic. On strong items it achieves $\geq 70\%$; on control items approximately 25\%, confirming a 40pp separation.

\paragraph{Benchmark integrity.} All distractors share the last 50\% of digits (arithmetic) or fall within $\pm 0.1$ decimal (fractions). Correct option position is balanced across A/B/C/D.

\section{Per-Category Radar Charts}
\label{sec:appendix_radar}

\begin{figure}[h]
    \centering
    \includegraphics[width=0.65\textwidth]{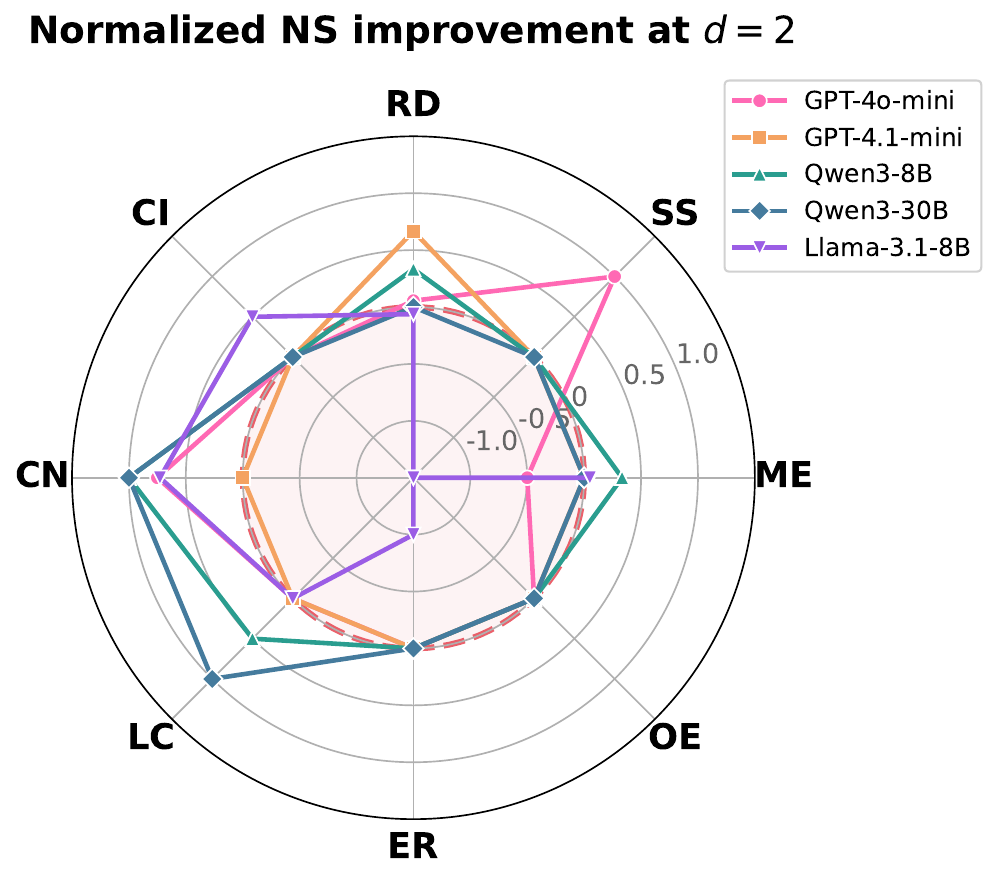}
    \caption{Normalized NS improvement at $d{=}2$.}
    \label{fig:radar_d2}
\end{figure}

\begin{figure}[h]
    \centering
    \includegraphics[width=0.65\textwidth]{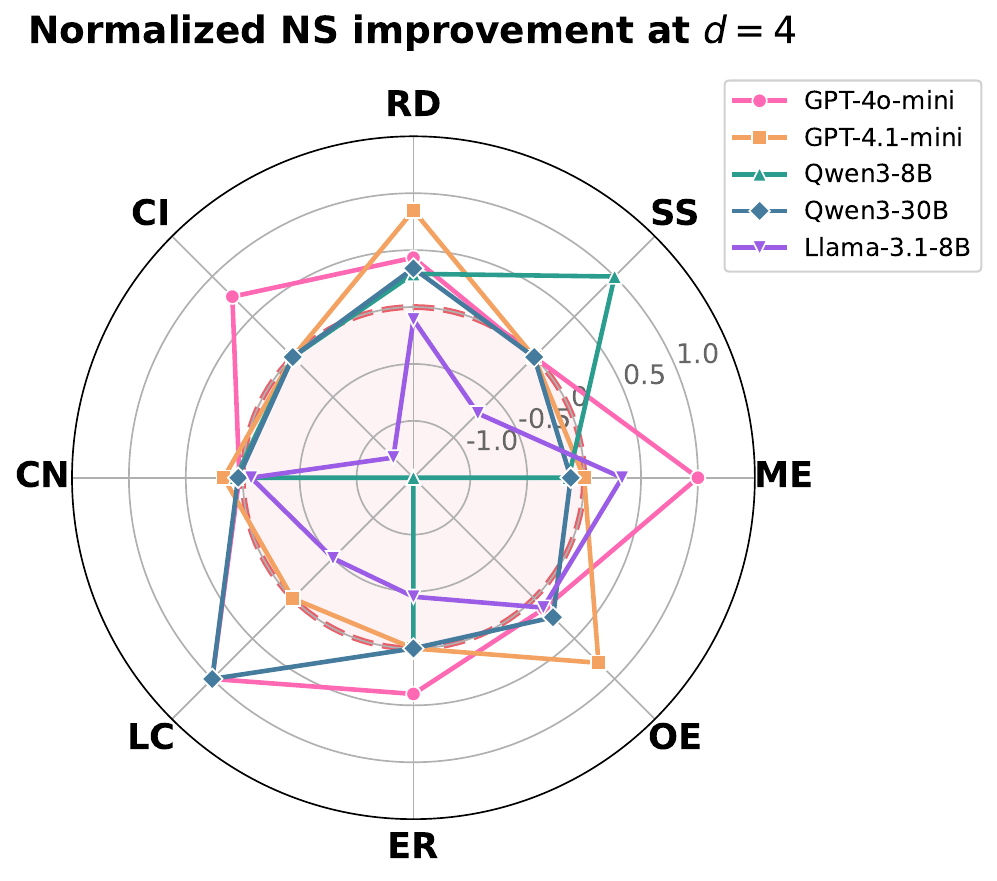}
    \caption{Normalized NS improvement at $d{=}4$.}
    \label{fig:radar_d4_app}
\end{figure}

\begin{figure}[h]
    \centering
    \includegraphics[width=0.65\textwidth]{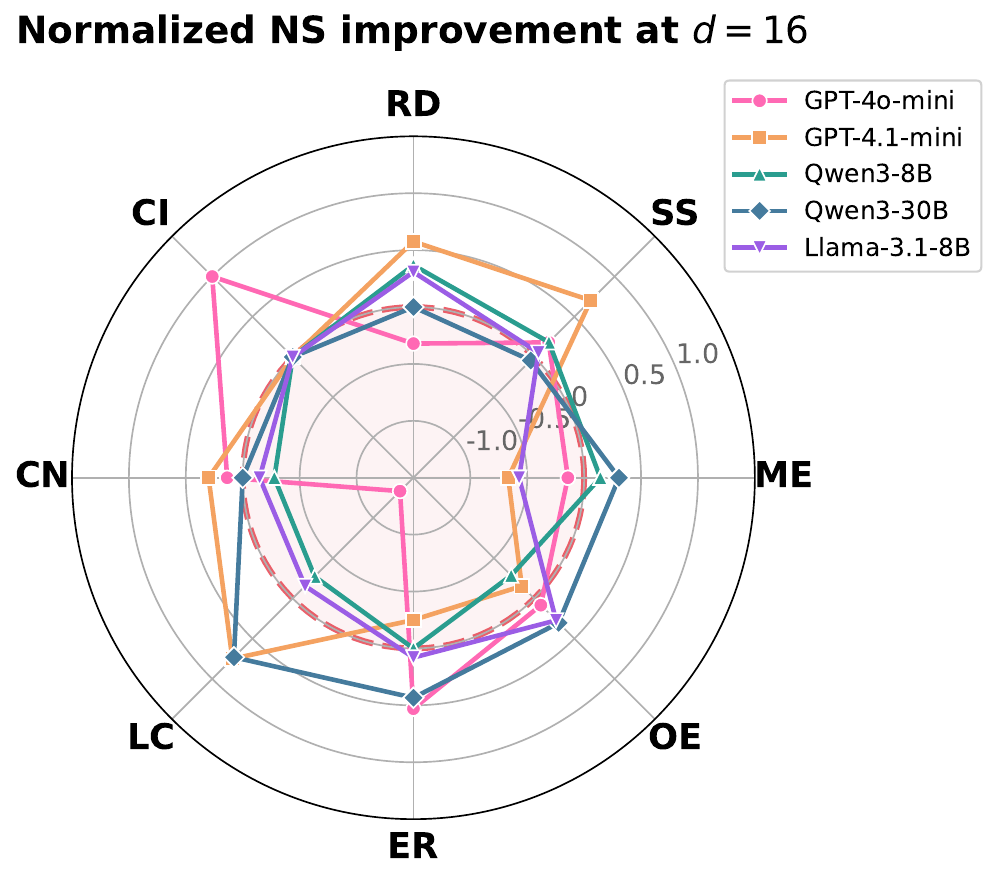}
    \caption{Normalized NS improvement at $d{=}16$.}
    \label{fig:radar_d16}
\end{figure}

Figures~\ref{fig:radar_d2}--\ref{fig:radar_d16} show the normalized NS improvement $(\text{NS}-\text{CoT})/(1-\text{CoT})$ for digit scales $d \in \{2, 4, 16\}$.

\section{Generate Task: Side-by-Side Model Outputs}
\label{sec:appendix_g_examples}

Figure~\ref{fig:g_examples} shows raw outputs from all five models on the same G prompt (magnitude\_estimation, $d{=}4$). Each model generates a strong-shortcut and a control problem pair. All 6 deterministic checks are shown for each model: {\color{green!50!black}\checkmark} = passed, {\color{red!70!black}\ding{55}} = failed.

\newcommand{\cmark}{{\color{green!50!black}\checkmark}}
\newcommand{\xmark}{{\color{red!70!black}\ding{55}}}

\begin{figure*}[h]
\small

\begin{tcolorbox}[width=0.9\linewidth, colback=green!5, colframe=green!50!black, title={\textbf{Qwen3-30B — 6/6 PASS}}, arc=2pt]
\textbf{Strong:} $10{,}200 \times 9{,}800 = 99{,}960{,}000$\\
{\color{blue}\textit{Reasoning:}} ``Both numbers are very close to 10{,}000, a power of 10, so rounding each to 10{,}000 and multiplying gives a very accurate estimate.''\\
\textbf{Control:} $4{,}321 \times 5{,}678 = 24{,}537{,}538$\\
\textit{Reasoning:} ``The numbers 4{,}321 and 5{,}678 are not close to any common power of 10.''\\[2pt]
\cmark\ S.Ans \quad \cmark\ C.Ans \quad \cmark\ SC.Exist \quad \cmark\ C.Block \quad \cmark\ Var \quad \cmark\ Novel
\end{tcolorbox}

\begin{tcolorbox}[width=0.9\linewidth, colback=red!5, colframe=red!70!black, title={\textbf{GPT-4.1-mini — 4/6 PASS}}, arc=2pt]
\textbf{Strong:} $1{,}980 \times 2{,}050$; model claims $= 4{,}000{,}000$, actual $= 4{,}059{,}000$\\
{\color{blue}\textit{Reasoning:}} ``Both numbers are close to 2{,}000, so rounding each to 2{,}000 gives a quick estimate.''\\
\textbf{Control:} $1{,}873 \times 2{,}147 = 4{,}019{,}531$\\[2pt]
\xmark\ S.Ans {\scriptsize(confused estimate with exact)} \quad \cmark\ C.Ans \quad \xmark\ SC.Exist {\scriptsize(2{,}000 $\neq$ power of 10)} \quad \cmark\ C.Block \quad \cmark\ Var \quad \cmark\ Novel
\end{tcolorbox}

\begin{tcolorbox}[width=0.9\linewidth, colback=red!5, colframe=red!70!black, title={\textbf{GPT-4o-mini — 5/6 PASS}}, arc=2pt]
\textbf{Strong:} $4{,}800 \times 2{,}100 = 10{,}080{,}000$\\
{\color{blue}\textit{Reasoning:}} ``Round both to 5{,}000 and 2{,}000, then multiply: $5{,}000 \times 2{,}000 = 10{,}000{,}000$.''\\
\textbf{Control:} $4{,}875 \times 2{,}137 = 10{,}420{,}075$\\[2pt]
\cmark\ S.Ans \quad \cmark\ C.Ans \quad \xmark\ SC.Exist {\scriptsize(rounding to 5k/2k, not power of 10)} \quad \cmark\ C.Block \quad \cmark\ Var \quad \cmark\ Novel
\end{tcolorbox}

\begin{tcolorbox}[width=0.9\linewidth, colback=red!5, colframe=red!70!black, title={\textbf{Qwen3-8B — 5/6 PASS}}, arc=2pt]
\textbf{Strong:} $2{,}500 \times 4{,}000 = 10{,}000{,}000$\\
{\color{blue}\textit{Reasoning:}} ``Round both to the nearest power of 10: $2{,}500 \approx 10{,}000$ and $4{,}000 \approx 10{,}000$.''\\
\textbf{Control:} $3{,}125 \times 7{,}680 = 24{,}000{,}000$\\[2pt]
\cmark\ S.Ans \quad \cmark\ C.Ans \quad \xmark\ SC.Exist {\scriptsize($2{,}500 \to 10{,}000$ is $4\times$ error; product is trivially exact)} \quad \cmark\ C.Block \quad \cmark\ Var \quad \cmark\ Novel
\end{tcolorbox}

\begin{tcolorbox}[width=0.9\linewidth, colback=red!5, colframe=red!70!black, title={\textbf{Llama-3.1-8B — 5/6 PASS}}, arc=2pt]
\textbf{Strong:} $8{,}500 \times 9{,}600 = 81{,}600{,}000$\\
{\color{blue}\textit{Reasoning:}} ``Round both to 10{,}000 and multiply.''\\
\textbf{Control:} $7{,}200 \times 8{,}900 = 64{,}120{,}000$ {\scriptsize(actual $= 64{,}080{,}000$)}\\[2pt]
\cmark\ S.Ans \quad \cmark\ C.Ans \quad \xmark\ SC.Exist {\scriptsize($8{,}500 \to 10{,}000$ gives 22\% error, too large)} \quad \cmark\ C.Block \quad \cmark\ Var \quad \cmark\ Novel
\end{tcolorbox}

\caption{G task: side-by-side model outputs for magnitude\_estimation ($d{=}4$). All 6 checks shown per model. Only Qwen3-30B passes all checks by choosing operands genuinely close to $10^4$. The dominant failure across all other models is \xmark\ SC.Exist: operands that \emph{look} round but do not satisfy the structural constraint for an effective power-of-10 shortcut.}
\label{fig:g_examples}
\end{figure*}

\section{Program-Based Generation}
\label{sec:appendix_generation}

All items are generated programmatically via category-specific Python generators with rejection sampling. This ensures:
(i) 100\% answer correctness by construction (exact arithmetic at generation time);
(ii) distractors that share the last 50\% of digits with the correct answer (arithmetic categories) or fall within $\pm 0.1$ decimal (fraction categories), preventing option-level elimination;
(iii) strong and control variants of each item differ only in numerical values, not template structure;
(iv) operands in control items that are ``hard numbers'' (last two digits in [25, 75], not divisible by 10, not near round boundaries), verified by automated checks.

Scaling operand size increases computational load while leaving shortcut applicability unchanged in strong variants.

\section{Full Results}
\label{sec:appendix_full_results}

Table~\ref{tab:full_results} presents accuracy and shortcut usage rate for all five models under all three prompting conditions (CoT, NS, Strict) across four digit scales.

\input{figures/TABLE_full_results.tex}

\section{Training Results}
\label{sec:appendix_training}

\paragraph{Experimental setup.}
We fine-tune Qwen3-8B and Llama-3.1-8B using LLaMA-Factory.
For both SFT and DPO variants, we use LoRA (rank 8, target = all linear layers) with the following shared hyperparameters: learning rate $5 \times 10^{-6}$, cosine scheduler with 10\% warmup, 3 epochs, cutoff length 2048, bf16 precision.
SFT uses batch size 2 with gradient accumulation 4 (effective batch size 8); DPO uses batch size 1 with gradient accumulation 8 (effective batch size 8), $\beta = 0.1$, sigmoid loss.

\paragraph{Training data.}
We construct training pairs from 500 MATH problems.
\emph{NS-DPO}: preferred responses are NS-style shortcut solutions (generated by GPT-5.1); rejected responses are verbose CoT solutions.
\emph{Baseline-DPO}: preferred = CoT solutions; rejected = NS solutions (reversed preference).
\emph{NS-SFT / CoT-SFT}: single-response fine-tuning on NS or CoT solutions respectively.
\emph{Combined-SFT}: fine-tuning on both NS and CoT solutions.
\emph{Combined-DPO}: DPO with NS preferred over CoT on the combined set.

\paragraph{Evaluation.}
In-domain evaluation uses a held-out 200-problem MATH set under both CoT and NS prompting. OOD evaluation covers 7 benchmarks (BBH, ARC, GPQA, MMLU-Pro, MedQA, LogiQA, CSQA) to verify no catastrophic forgetting.

Tables~\ref{tab:training} and~\ref{tab:ood} present the full results.

\input{figures/TABLE_training.tex}
\input{figures/TABLE_ood.tex}

\section{Strict Condition Analysis}
\label{sec:appendix_strict}

The Strict condition provides a negative control by explicitly forbidding shortcuts (full data in Table~\ref{tab:full_results}).
Open-weight models frequently truncate responses at the 512-token limit (30--60\% of items), deflating accuracy estimates.
At $d{=}4$, Strict achieves 90.5\% on strong items for GPT-4o-mini (vs.\ NS 84.5\%), but Strict control accuracy (87.2\%) far exceeds NS control (63.0\%).
Shortcut-related keywords appear in only 11\% of Strict responses, compared to 46\% under CoT and 87\% under NS, confirming the gradient Strict $<$ CoT $<$ NS.

\section{CoT Strategy Analysis}
\label{sec:appendix_cot_strategy}

CoT does not uniformly produce pure computation.
On cancellation items (e.g., $3{,}456 + 7{,}891 - 7{,}889$), GPT-4o-mini's CoT response discovers ``$7{,}891 - 7{,}889 = 2$'' mid-solution.
On structural-shortcut items, Qwen3-30B frequently invokes the distributive law (``$99 \times 37 = (100-1) \times 37$'') unprompted.
This spontaneous shortcut discovery means the NS--CoT comparison is \emph{conservative}: CoT already activates some shortcuts on its own, yet the NS prompt still produces measurable additional asymmetric effects.

The SU rate rows in Table~\ref{tab:full_results} quantify this: under CoT, shortcut rates are 20--39\% at $d{=}4$, with a consistent strong$>$control gap.
NS prompting raises shortcut use to 68--86\% uniformly across strong and control items (e.g., Qwen3-8B: 86\% strong, 88\% control at $d{=}4$), revealing the mechanism behind the accuracy asymmetry: NS induces shortcuts uniformly, but they succeed only where valid shortcuts exist.

\section{Human Validation of the SU Rate Judge}
\label{sec:appendix_human_validation}

The SU rate metric in Table~\ref{tab:full_results} relies on GPT-4.1-mini as an automated judge to classify each model response as SHORTCUT or COMPUTATION.
To validate this judge, we conducted a human annotation study on 100 randomly sampled responses from $d{=}4$, stratified by model (20 per model), prompting condition (CoT vs.\ NS), and variant (strong vs.\ control).

\paragraph{Annotation protocol.}
One annotator labeled each response as SHORTCUT or COMPUTATION based on the following criterion: a response counts as SHORTCUT if it exploits proximity to round numbers, benchmark comparison, near-cancellation, or structural decomposition to simplify computation; standard partial-product expansion (e.g., $4386 = 4000 + 300 + 80 + 6$) counts as COMPUTATION.

\paragraph{Results.}
The human shortcut rates closely track the automated judge:
under CoT, the human annotator labels 40\% of responses as SHORTCUT (judge: 20--39\%);
under NS, 88\% (judge: 68--86\%).
Across all 20 model$\times$condition$\times$variant cells, the Pearson correlation between human and judge shortcut rates is $r = 0.85$.

The human rates are systematically higher than the judge rates (mean $|\Delta| = 18.7$pp), indicating that the judge applies a \emph{stricter} classification threshold.
Disagreements fall into two main categories:
(1)~\emph{Estimation misclassified as computation}: responses that round operands to non-power-of-10 values (e.g., $4770 \to 4800$, $3207 \to 3200$) are sometimes classified as COMPUTATION by the judge despite using estimation—the judge reserves SHORTCUT for power-of-10 rounding, while humans accept broader rounding strategies.
(2)~\emph{Landmark computation}: for landmark-comparison items (e.g., ``is 25\% of 51340 $>$ 17242?''), models compute $25\% = \div 4$ exactly rather than using a benchmark shortcut; both human and judge agree these are COMPUTATION, but they inflate the overall CoT shortcut rate because the category itself does not clearly separate estimation from exact fractional arithmetic.
This means the SU rates reported in Table~\ref{tab:full_results} are \emph{conservative lower bounds}, and the core finding---that NS prompting substantially increases shortcut usage---holds under both human and automated classification.

\end{document}

%% file: sections/0_abstract.tex
Large language models often default to step-by-step computation even when efficient numerical shortcuts are available. This raises a basic question: do they exhibit \emph{number sense} in a human-like behavioral sense, i.e.,  the ability to recognize numerical structure, apply shortcuts when appropriate, and avoid them when they are not?
We introduce \textsc{SenseMath}, a controlled benchmark for evaluating structure-sensitive numerical reasoning in LLMs. \textsc{SenseMath} contains 4{,}800 items spanning eight shortcut categories and four digit scales, with matched strong-shortcut, weak-shortcut, and control variants. It supports three evaluation settings of increasing cognitive demand: \emph{Shortcut Use} (whether models can apply shortcuts on shortcut-amenable problems); \emph{Applicability Judgment} (whether they can recognize when a shortcut is appropriate or misleading); and \emph{Problem Generation} (whether they can generate new problem items that correctly admit a given type of shortcut).
Our evaluation across five LLMs, ranging from GPT-4o-mini to Llama-3.1-8B, shows a consistent pattern: {when explicitly prompted, models readily adopt shortcut strategies and achieve substantial accuracy gains on shortcut-amenable items (up to 15\%), yet under standard chain-of-thought prompting they spontaneously employ such strategies in fewer than 40\% of cases, even when they demonstrably possess the requisite capability. Moreover, 
models systematically \emph{over-generalise} shortcuts to problems where they do \emph{not apply}, and fail to \emph{generate} valid shortcut-bearing problems from scratch. Together, these results suggest that current LLMs exhibit \emph{procedural} shortcut fluency without the \emph{structural} understanding of when and why shortcuts work that underlies human number sense.}
Our code and data are available at \url{https://github.com/zhmzm/SenseMath}.


%% file: sections/1_introduction.tex
\section{Introduction}
\label{sec:introduction}
\vspace{-0.05in}
Recent advances in large language models (LLMs) have led to remarkable progress on arithmetic and mathematical reasoning tasks
Chain-of-thought (CoT) prompting \citep{wei2022chain, kojima2022large} and reinforcement-learning-based approaches such as OpenAI o1 \citep{openai2024o1} and DeepSeek-R1 \citep{guo2025deepseek} have pushed performance on mathematical benchmarks (e.g., GSM8K \citep{cobbe2021training} and MATH \citep{hendrycks2021measuring}) to near-human levels.
However, it remains unclear whether this success reflects a genuine understanding of numerical structure. 

A central concept capturing  human mathematical reasoning ability is \emph{number sense}: the capacity to perceive structure in numbers, flexibly choose solution strategies (e.g., efficient computational strategies), and generalize them across contexts \citep{mcintosh1992proposed}. For example, when computing $98 \times 14$, 
a human can exploit proximity to 100 and compute $(100 - 2) \times 14$, whereas for $73 \times 14$, no such shortcut naturally applies and a different method is used. 
Likewise, when comparing  10/11 and   11/12, a human may avoid cross-multiplication altogether by observing that both fractions are close to 1 and that  11/12 is larger because it has the smaller gap to 1.


In developing and evaluating number sense, it is important to assess not only whether learners can \emph{apply} number sense strategies, but also whether they can engage with them at \emph{higher cognitive levels}. Following Bloom’s Taxonomy \citep{bloom1956taxonomy,krathwohl2002revision}, a widely used framework in education for assessing different levels of cognitive engagement, these higher  levels include  \emph{analysis, evaluation}, and \emph{creation}. Learners should be able to judge when a number sense strategy is appropriate, evaluate whether it is being used in problem solving, and generate new problems that meaningfully involve number sense. These abilities reflect deeper and more advanced understanding.
This view is consistent with research that frames number sense as a flexible, adaptive form of mathematical reasoning rather than a fixed set of procedures \citep{boaler2022mathematical,devlin2010mathematical}, highlighting the importance of how learners select, justify, and generate strategies in context.

Despite their importance, current evaluations of LLM reasoning do not directly test these capabilities. Most benchmarks focus on answer correctness \citep{cobbe2021training, hendrycks2021measuring, lightman2024lets, aime2024}, while studies of CoT plausibility examine whether reasoning traces faithfully reflect the model's internal computation \citep{turpin2024language, lanham2023measuring, paul2024making}.
While useful, these metrics primarily assess whether a model can execute a reasoning procedure, and do not distinguish between \emph{procedural competence} and a \emph{deeper, declarative understanding} of numerical structure.

 
This limitation is particularly relevant in \emph{educational} settings, where LLMs are increasingly used as AI tutors. Effective teaching requires more than demonstrating how to solve a problem: it involves helping students understand when a strategy is applicable, when it is not,  how to recognize the underlying structure, and generating new  problems for students to practice these distinctions. 
  
Motivated by such gaps in current evaluations of mathematical reasoning, we study whether LLMs exhibit behaviors consistent with number sense through three progressively more demanding capabilities, formalized as the following research questions: 
\vspace{-0.05in}
\begin{tcolorbox}[colback=gray!10, colframe=gray!50, arc=2pt, boxrule=0.5pt, left=6pt, right=6pt, top=3pt, bottom=3pt]
\begin{itemize}[leftmargin=*,itemsep=2pt] \small
    \item  RQ1: Do LLMs invoke such shortcuts spontaneously in the absence of explicit instruction?
    \item   RQ2: In cases where shortcuts are not invoked spontaneously, can LLMs apply them when explicitly instructed?
    \item RQ3:  Do LLMs misuse shortcuts when they appear applicable but are not?
    \item RQ4: Can LLMs generate new problems that admit a given type of shortcut?
\end{itemize}
\end{tcolorbox}\vspace{-0.05in}
These RQs correspond to increasing cognitive demands: from \emph{\textbf{applying}} a known method (RQ1,RQ2), to \emph{\textbf{judging}} when it should be used (RQ3), to \emph{\textbf{creating}} new items (RQ4). Despite its importance, evaluating number sense in LLMs is inherently challenging. Unlike standard mathematical reasoning, number sense is not a single well-defined skill, but rather a flexible meta-cognitive disposition  \citep{sowder1992making}. As a result, valid assessment requires carefully constructed instruments that go beyond accuracy on isolated tasks and instead probe behavior across controlled variations \citep{kirkland2024validity}.
Existing efforts provide only a partial view of this capability.
While LLMs can solve complex reasoning problems, they often fail on basic numerical operations such as magnitude comparison and digit manipulation \citep{yang2024number}. Prior evaluations primarily target narrow, perception-level aspects of numerical ability: for instance, \citet{testolin2024can} study elementary numerical discrimination, and \citet{rahman2025fragile} evaluate combinatorial reasoning over numbers using tasks such as the Game of 24. Although informative, these tasks largely assess whether a model can \emph{perceive} numerical properties, rather than whether it can \emph{reason with them}, in particular, whether it can selectively apply efficient strategies when the problem structure permits.


To bridge this gap, we introduce \textsc{SenseMath}, a controlled benchmark designed to evaluate structure-sensitive numerical reasoning. The benchmark consists of 4{,}800 items spanning eight categories and four digit scales ($d \in {2,4,8,16}$), covering both problem-level shortcuts (e.g., magnitude estimation, structural decomposition, relative distance, cancellation, compatible numbers, and landmark comparison), reasoning-level shortcuts (equation-based transformations), and option-level shortcuts (e.g., elimination strategies). Each item has matched strong-shortcut, weak-shortcut, and control variants, enabling causal attribution of performance differences to strategy selection rather than underlying problem difficulty. This design allows us to systematically test whether models exploit available structure, avoid overgeneralizing shortcuts, and remain stable when such structure is absent. An overview of the benchmark is shown in Figure~\ref{fig:overview}.

\vspace{-0.1in}
\paragraph{Key findings.}
We evaluate five models (GPT-4o-mini, GPT-4.1-mini, Qwen3-30B, Qwen3-8B, and Llama-3.1-8B) under varying prompting conditions, over a total of 72{,}000 inferences  on \textsc{SenseMath}. 
These evaluations span three levels of increasing cognitive demand: \emph{Shortcut Use}, \emph{Applicability Judgment}, and \emph{Problem Generation}. Our findings map directly to the research questions above. \vspace{-0.05in}
{\begin{itemize}[leftmargin=*,itemsep=2pt]
    \item \textbf{RQ1: Sometimes.} Under standard CoT prompting, shortcut strategies appear in fewer than 40\% of responses at $d{=}4$, though this rate increases with digit scale.
    \item \textbf{RQ2: Yes, but more for capable models.} When explicitly instructed, capable models achieve accuracy gains of up to 15\% (GPT-4.1-mini at $d{=}8$), while 8B-parameter models show no benefit or even degradation.
    \item \textbf{RQ3: Yes, severely misused.} Models accept nearly all problems as shortcut-amenable (control rejection rate as low as 0\%) and over-apply estimation strategies, reducing accuracy on control items by up to 12\%.
    \item \textbf{RQ4: Largely failed in generation.} Models construct valid shortcut-amenable problems at only 2--24\% pass rate, reproducing the surface form but not the structural constraints that make a shortcut effective.
\end{itemize}}



\vspace{-0.1in}
\paragraph{Contributions.} 
This work makes the following contributions:\vspace{-0.05in}
\begin{itemize}[leftmargin=*,itemsep=2pt]
\item We introduce a new perspective for evaluating numerical reasoning in LLMs through the lens of number sense, organizing evaluation by the level of cognitive engagement following Bloom’s revised taxonomy \citep{krathwohl2002revision}, spanning Apply (\emph{Shortcut Use}), Analyze (\emph{Applicability Judgment}), and Create (\emph{Problem Generation}).
\item  We instantiate this framework with \textsc{SenseMath}, a controlled benchmark of 4{,}800 items designed to isolate structure-sensitive behavior. Its construction is guided by cognitive load theory \citep{sweller1988cognitive}: items vary along two orthogonal dimensions of intrinsic cognitive load:  (1) number of digits and (2) number-sense strategy applicability,  while extraneous load is held constant through a shared surface template \citep{ober2023development}.
\item Empirically, we show that some LLMs can execute shortcuts when prompted but most of them struggle to use them appropriately or generate valid new items.
\end{itemize}

\textbf{Impact:}  
Our results reveal a gap between procedural execution and structural numerical understanding in LLMs: models may be able to execute shortcut-based solutions when explicitly prompted, but are substantially less reliable at judging when such strategies apply or at generating new examples. Since these abilities are important for supporting flexible mathematical reasoning, our findings suggest caution when deploying LLMs as autonomous math tutors. 

%% file: sections/2_related_work.tex
\vspace{-0.03in}
\section{Related Work}
\label{sec:related}\vspace{-0.06in}

\textbf{Number sense and its assessment in LLMs.}
The concept of number sense originates in mathematics education: \citet{mcintosh1992proposed} define it as a person's general understanding of number and operations together with the inclination to use this understanding flexibly, while \citet{mcintosh1992proposed} identify core components such as magnitude estimation, benchmark use, and recognition of numerical structure.
\citet{sowder1992making} further characterise number sense as a meta-cognitive disposition to choose efficient computational paths, and \citet{yang2003assessing} operationalise these ideas into validated instruments for school-age children.
Recent work has begun probing analogous abilities in LLMs. \citet{yang2024number} show that models excel at complex reasoning yet fail on basic numerical operations like magnitude comparison and digit manipulation. \citet{rahman2025fragile} find strong deterministic performance but failure on heuristic-search tasks such as the Game of 24, suggesting a ``fragile number sense.'' \citet{li2025numeracy} expose systematic gaps across six numeracy capabilities, and \citet{nikankin2024arithmetic} reveal that LLMs solve arithmetic via sparse neuron-level heuristics rather than learned algorithms.
However, these efforts probe only narrow, perception-level facets of number sense. \textsc{SenseMath} goes further by testing whether models can \emph{selectively activate} shortcut strategies through prompting, and evaluates number sense at three cognitive levels (\emph{Apply - Shortcut Use, Analyze - Applicability Judgment}, and  \emph{Create - Problem Generation}). 

\textbf{LLM reasoning and evaluation.}
Chain-of-thought prompting \citep{wei2022chain, kojima2022large} and its variants \citep{wang2023selfconsistency, yao2023tree} have substantially improved LLM performance on mathematical benchmarks such as GSM8K \citep{cobbe2021training}, MATH \citep{hendrycks2021measuring}, MathBench \citep{liu2024mathbench}, and MathVista \citep{lu2024mathvista}.
Meanwhile, a parallel line of work questions the reliability of these reasoning traces: \citet{turpin2024language} demonstrate that CoT explanations can be systematically unfaithful, and diagnostic benchmarks such as GSM-Symbolic \citep{mirzadeh2024gsmsymbolic} and MATH-Perturb \citep{huang2025mathperturb} expose fragilities behind high aggregate scores.
On the efficiency side, methods like Chain of Draft \citep{kim2024chain}, Sketch-of-Thought \citep{aytes2025sketch}, and Token-Budget-Aware reasoning \citep{han2024tokenbudget} aim to compress verbose reasoning, while \citet{chen2024overthinking} show that O1-like models waste computation on trivial problems.
These approaches share a common assumption: the underlying strategy is fixed, and the goal is to express it more concisely. By contrast, number sense involves switching to a qualitatively \emph{different strategy}  that exploits numerical structure to bypass unnecessary computation.   Our work is therefore complementary: existing methods compress the \emph{same} strategy; \textsc{SenseMath} measures if models can 
\emph{recognize when a different shortcut-based strategy is available and use it}.

%% file: sections/3_benchmark.tex
\vspace{-0.05in}
\section{The \textsc{SenseMath} Benchmark}
\label{sec:benchmark}\vspace{-0.05in}
This section describes our design philosophy, category taxonomy, programmatic generation pipeline, and evaluation framework.

\subsection{Design Philosophy} \vspace{-0.05in}
\label{sec:design}
We design our evaluation around how humans exhibit \emph{number sense} in arithmetic: not merely by producing correct answers, but by recognizing numerical structure,  selecting efficient
shortcuts when appropriate,   avoiding them when they are not, and   generating new examples for further  practice. 
Our goal is therefore not to test whether LLMs can compute numerical answers by any means available, but whether they can reason in a way that reflects structure-sensitive shortcut use. We do not argue that LLMs should always solve math problems in this way. In many applications, models can directly obtain correct answers by calling external tools such as calculators or programs.
That, however, is not the focus of this work. Instead, we study number sense from an educational perspective: whether LLMs can recognize when a shortcut applies, explain it clearly, avoid misusing it, and generate new examples for students to practice.


\begin{figure}[t]
    \centering
    \includegraphics[width=0.72\textwidth]{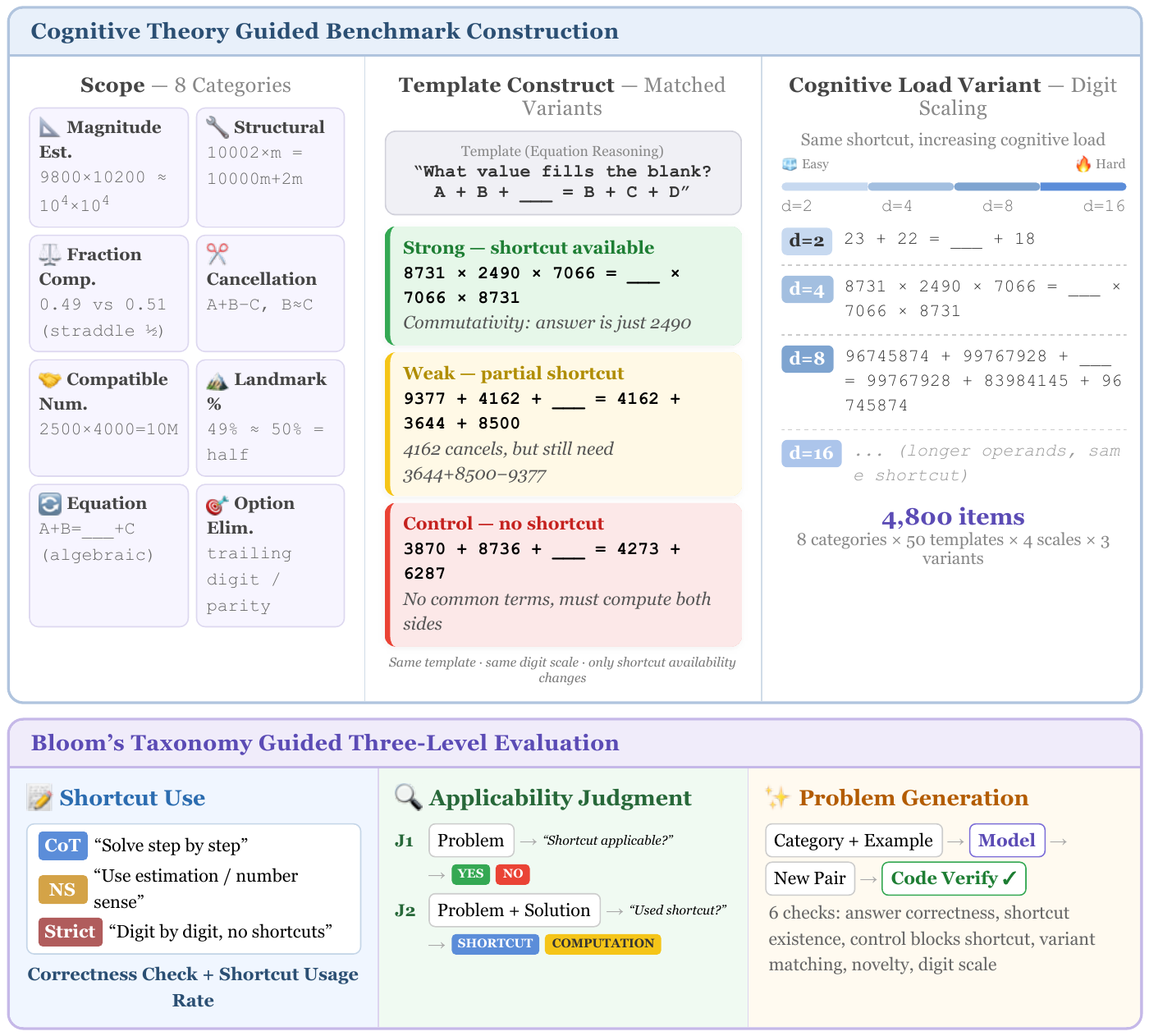} \vspace{-0.1in}
    \caption{\textsc{SenseMath} overview. Each item has matched strong-shortcut, weak-shortcut, and control variants administered under CoT and number-sense prompting. The matched design isolates selective shortcut exploitation from general prompt effects. }
    \label{fig:overview} \vspace{-0.2in}
\end{figure}

\textbf{Shortcut} is a reasoning strategy that exploits numerical structure to simplify or bypass standard step-by-step computation. Examples include using proximity to base values, cancellation, or relative comparisons, such as rewriting $98 \times 14$, as  $ (100-2)  \times 14$,   or comparing  10/11 and  11/12 by noting that both are close to 1 and comparing their gaps to 1. A transformation does not qualify as a shortcut if the simplified form still requires essentially the same level of computation,  for example, rounding $5,374  \times 2,169$   to $5,400  \times 2,200$   still leaves a multi-digit multiplication.

We follow Cognitive Load Theory \citep{sweller1988cognitive} to construct items with shortcut-based solutions and controlled variants. We hold extraneous load constant through 50 uniform surface templates and a fixed multiple-choice format, while varying intrinsic load along two orthogonal dimensions:  digit scale ($d \in \{2,4,8,16\}$) and shortcut availability (strong/weak/control), described next. 

\textbf{Shortcut-Invariant Scaling.} 
Given a question template that admits a particular shortcut, we can generate multiple items by varying the digit scale while preserving the same underlying structural cue. This allows us to test whether models can maintain shortcut-based reasoning as surface numerical complexity increases. Importantly, the digit scale is designed to be \emph{shortcut-invariant}: although the numbers become longer, the structural relation that licenses the shortcut remains unchanged. For example, 
the reasoning shortcut used to compare  10/11 and   11/12 is identical to that used for  comparing   1110/1111 and  1111/1112 (comparing their gaps to 1), despite the larger number of digits.

\textbf{Matched Shortcut Variants.}
For each problem template at a fixed digit scale, we construct \textbf{three matched variants}: \emph{\textbf{strong}}, where a clean and effective shortcut is available; \emph{\textbf{weak}}, where a shortcut is partially helpful but still requires additional computation; and \emph{\textbf{control}}, where no effective shortcut applies. These variants share the same template and digit scale and differ only in their numerical instantiation, ensuring that surface form and overall problem difficulty remain comparable. This matched design allows us to isolate the effect of shortcut availability and attribute performance differences to strategy selection rather than confounding factors.



\vspace{-0.05in}
\subsection{Categories of Shortcuts}
\label{sec:categories}\vspace{-0.05in}

\textsc{SenseMath} includes 8 categories of shortcuts, organized into three tiers based on where the shortcut operates.  These categories cover core forms of number-sense reasoning   competencies identified in the mathematics education literature \citep{mcintosh1992proposed}.

\textbf{Tier 1: Problem-level shortcuts} \vspace{-0.1in}
\begin{enumerate}[leftmargin=*,itemsep=1pt]
    \item \emph{Magnitude estimation (ME)} tests the ability to approximate products by rounding operands to nearby powers of 10, which is   foundational skill in gauging whether an answer is ``in the right ballpark.''  \vspace{-0.03in}
    \item \emph{Structural shortcuts (SS)} require recognising and exploiting algebraic identities near round numbers (e.g., $99 \times 37 = (100{-}1) \times 37$), probing whether models can decompose computations into simpler parts. \vspace{-0.03in}
    \item \emph{Relative distance (RD)} measures relational reasoning about fractions by comparing their distance from a common benchmark (e.g., $\nicefrac{1}{2}$), avoiding full cross-multiplication. \vspace{-0.03in}
    \item \emph{Cancellation (CI)} tests sensitivity to near-cancellation patterns in expressions like $A + B - C$ when $B \approx C$, reducing a three-operand computation to simple subtraction. \vspace{-0.03in}
    \item \emph{Compatible numbers (CN)} assesses whether models can identify product-friendly rounding opportunities (e.g., recognising that $248 \times 4{,}012 \approx 250 \times 4{,}000$), a key mental-math strategy. \vspace{-0.03in}
    \item \emph{Landmark comparison (LC)} probes the use of familiar reference points when comparing percentages or fractions (e.g., $49\% \approx 50\%$), reflecting the human tendency to anchor judgments to well-known values.
\end{enumerate}\vspace{-0.05in}

\textbf{Tier 2: Reasoning-level shortcut.}
\emph{Equation reasoning (ER)} moves beyond arithmetic to algebraic structure: recognising identities such as commutativity and common-term cancellation in fill-in-the-blank equations, reducing multi-step algebra to a single structural observation.

\textbf{Tier 3: Option-level shortcut.}
\emph{Option elimination (OE)} tests a meta-reasoning ability: ruling out answer choices by quick feasibility checks (trailing digit, parity, order-of-magnitude) without computing the exact answer, reflecting strategic use of  multiple-choice format itself.

For each category, we generate 50 question templates. Each templates is instantiated at four digit scales ($d \in \{2,4,8,16\}$) through \emph{Shortcut-Invariant Scaling}, and each scaled item is further expanded into three \emph{Matched Shortcut Variants} (\emph{strong}, \emph{weak}, and \emph{control}). In total, this yields $8 \times 50 \times 4 \times 3 = 4{,}800$ problem items.
Example items for each category are provided in Appendix~\ref{sec:appendix_examples}.
All items are generated programmatically via category-specific Python generators with rejection sampling to guarantee answer correctness and controlled distractor difficulty. Full generation details are provided in Appendix~\ref{sec:appendix_generation}.
\vspace{-0.1in}
\subsection{Three-Level Evaluation}
\label{sec:conditions}\vspace{-0.1in}

\textbf{Usage of Number Sense Strategy.}
To answer RQ1 and RQ2, we evaluate models under three prompting conditions that vary in the extent to which they encourage shortcut-based reasoning. \textbf{CoT} (chain-of-thought) uses a standard step-by-step reasoning prompt, testing whether models invoke shortcuts spontaneously. \textbf{NS} (number-sense) encourages mathematical intuition and easy calculations, but does not mention any specific shortcut type, making it deliberately category-agnostic. \textbf{Strict} explicitly forbids shortcuts and requires fully explicit computation, serving as a negative control.
All conditions require answers inside \texttt{\textbackslash boxed\{\}} for unambiguous extraction. Full prompt templates are provided in Appendix~\ref{sec:appendix_prompts}.
We use two metrics to report model performance: \textbf{Accuracy}, which measures final-answer correctness, and \textbf{Shortcut Usage Rate}, which measures the proportion of responses that use shortcut-based reasoning on instances where such strategies are applicable.





\textbf{Applicability Judgment.}
To evaluate whether models can distinguish between cases where shortcuts are appropriate and where they are not, we introduce two judgment tasks. \textbf{J1} presents a problem and asks whether it can be solved faster with shortcut reasoning (\texttt{YES}/\texttt{NO}). \textbf{J2} presents a problem together with a  solution and asks whether the solution uses a shortcut or a standard computation strategy (\texttt{SHORTCUT}/\texttt{COMPUTATION}). J1 measures \emph{recognition} of shortcut applicability, while J2 measures \emph{identification} of the strategy type used.

\textbf{Problem Generation.}
To test whether models can construct new items that admit a given shortcut, we ask each model, given a category description and one example, to generate a new \emph{strong}/\emph{control} problem pair. Generated items are verified using six deterministic code checks: answer correctness, shortcut existence, control blocking, variant matching, novelty, and digit scale consistency. This evaluates whether models can \emph{construct} valid shortcut-amenable problems with the intended structural properties.





Full task specifications and verification details are provided in Appendix~\ref{sec:appendix_tasks}.

%% file: sections/4_experiments.tex
\vspace{-0.05in}
\section{Experiments} 
\label{sec:experiments}\vspace{-0.1in}

We organise our experiments around the research questions posed in the introduction:
\textbf{RQ1:} Can LLMs invoke shortcuts spontaneously? (\S\ref{sec:rq1})
\textbf{RQ2:} Can LLMs apply shortcuts when explicitly instructed? (\S\ref{sec:rq2})
\textbf{RQ3:} Do LLMs misuse shortcuts when they are not appropriate? (\S\ref{sec:rq3})
\textbf{RQ4:} Can LLMs generate new shortcut-amenable problems? (\S\ref{sec:rq4})
We additionally investigate: \textbf{RQ5:} What benefits could NS post-training bring? (\S\ref{sec:rq5})

\vspace{-0.1in}
\subsection{Experimental Setup}
\label{sec:exp_setup} \vspace{-0.1in}

\paragraph{Models.}
We evaluate five instruction-tuned models spanning three model families:
Qwen3-30B and Qwen3-8B \citep{yang2025qwen3},
Llama-3.1-8B-Instruct \citep{grattafiori2024llama},
and GPT-4o-mini and GPT-4.1-mini \citep{openai2024gpt4o}.
GPT-4.1-mini is a more capable successor to GPT-4o-mini that achieves near-ceiling CoT accuracy on \textsc{SenseMath} at $d{=}4$, providing an important reference point for how strong baseline performance constrains the prompting asymmetry.
Open-weight models are served via vLLM with tensor parallelism; GPT models are accessed through the OpenAI API.
All inferences use greedy decoding (temperature$\,{=}\,0$) and $\texttt{max\_tokens}{=}512$.


\vspace{-0.1in}
\subsection{RQ1: Can LLMs Invoke Shortcuts Spontaneously?}
\label{sec:rq1} \vspace{-0.1in}

\begin{figure}[t]
    \centering
    \begin{minipage}[t]{0.55\textwidth}
        \centering
        \includegraphics[width=0.9\textwidth]{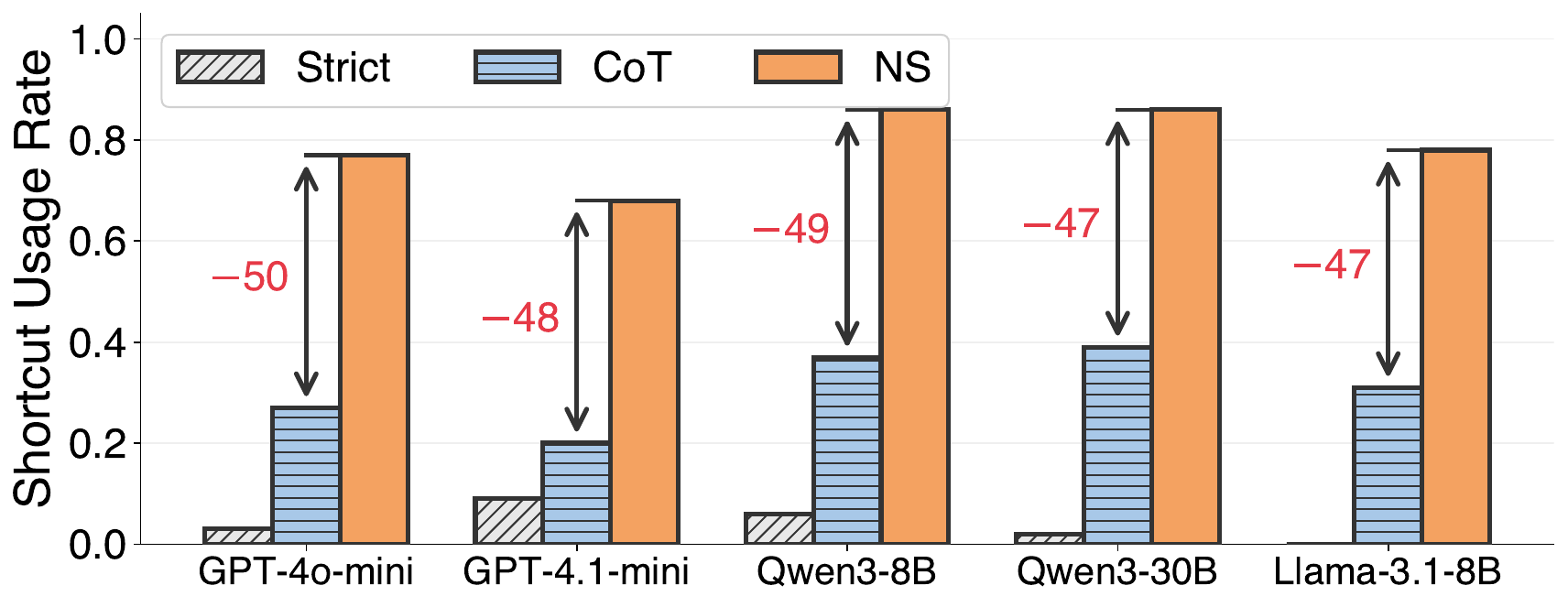}
        \caption{SU rate on strong items at $d{=}4$ across three prompting conditions.}
        \label{fig:su_bar_d4}
    \end{minipage}
    \hfill
    \begin{minipage}[t]{0.4\textwidth}
        \centering
        \includegraphics[width=0.8\textwidth]{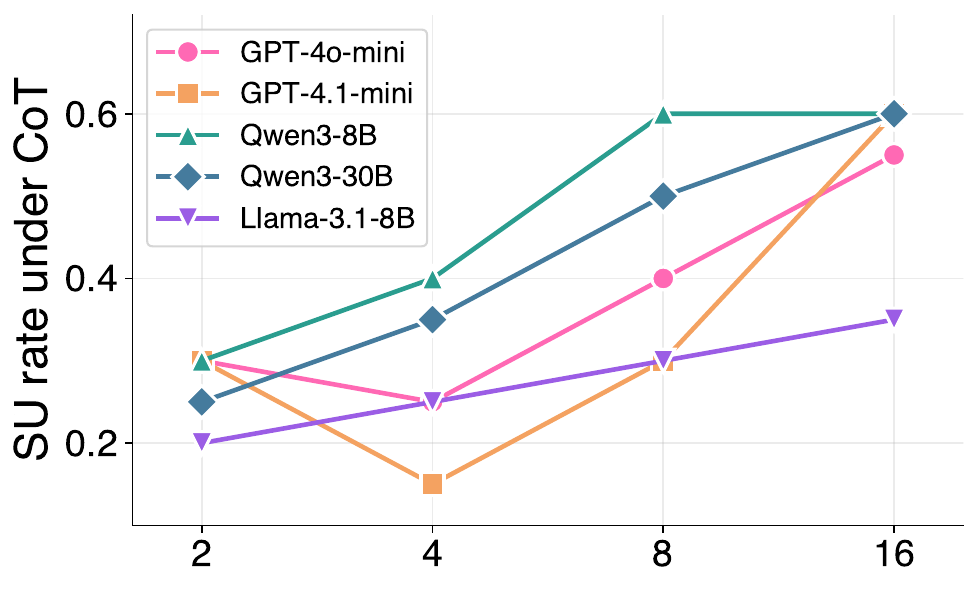}
        \caption{SU rate under CoT increases with digit scale.}
        \label{fig:su_cot_line}
    \end{minipage}
\end{figure}

\paragraph{CoT rarely triggers shortcuts.}
Figure~\ref{fig:su_bar_d4} shows the shortcut usage (SU) rate on strong-shortcut items at $d{=}4$ under all three prompting conditions.
Strict suppresses shortcuts almost entirely (SU $\leq$ 0.09), confirming it works as a negative control.
Under CoT, shortcuts appear in only 20--39\% of responses despite being available—models default to step-by-step computation even when efficient alternatives exist.
NS prompting dramatically raises SU to 68--86\% (red arrows in Figure~\ref{fig:su_bar_d4}), a $+$47--50pp increase across all models.
This gap reveals that number-sense reasoning is \emph{latent} in these models but requires explicit prompting to activate.
Furthermore, the SU rate under CoT increases with digit scale (Figure~\ref{fig:su_cot_line}): at $d{=}16$, even CoT elicits shortcuts in 37--59\% of responses, suggesting that as problems become harder, models increasingly discover shortcuts on their own.

\vspace{-0.05in}
\paragraph{NS benefit increases with problem difficulty.}
Figure~\ref{fig:rq1_combined}(b) shows that the accuracy gain from NS prompting generally increases with digit scale for capable models.
GPT-4.1-mini gains $+$1pp at $d{=}2$ but $+$15pp at $d{=}8$; Qwen3-30B shows a similar trajectory ($+$4 to $+$14pp).
This is because larger digit scales make brute-force computation increasingly error-prone under CoT, while shortcut strategies remain equally effective on strong-shortcut items regardless of operand size.
Notably, Llama-3.1-8B and Qwen3-8B do not follow this pattern: their NS gains remain near zero or negative at all scales, further confirming that 8B-parameter models lack robust number-sense capabilities.


\vspace{-0.05in}
\subsection{RQ2: Can LLMs Apply Shortcuts When Instructed?}
\label{sec:rq2}
\begin{figure}[t]
    \centering
    \includegraphics[width=0.48\columnwidth]{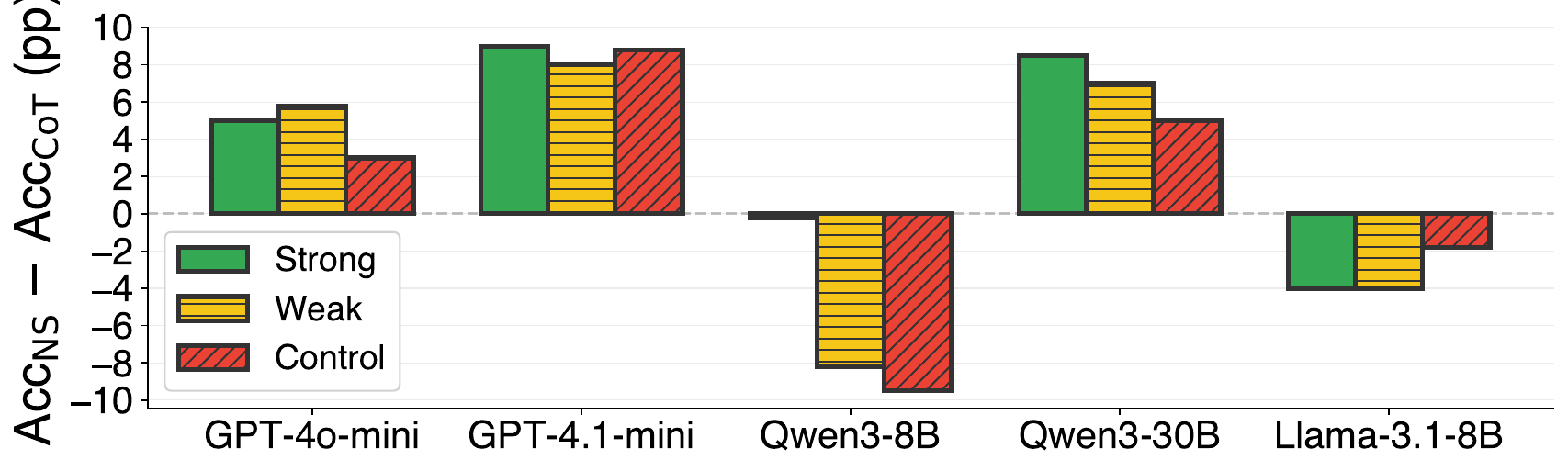}
    \caption{Average accuracy gain from NS prompting (vs.\ CoT) across all digit scales, by variant.}
    \label{fig:ns_gain_swc} \vspace{-0.1in}
\end{figure}

\vspace{-0.05in}
\paragraph{More  capable models use NS prompting effectively.}
Figure~\ref{fig:ns_gain_swc} reveals a clear asymmetry in how NS prompting affects different models and matched variants (strong/weak/control).
GPT-4.1-mini benefits uniformly across all three variants ($+$9pp on strong, $+$8pp on weak, $+$9pp on control), suggesting it genuinely improves its reasoning rather than blindly applying shortcuts.
GPT-4o-mini and Qwen3-30B show a similar but more selective pattern: strong items gain $+$5 to $+$9pp, while control items gain less ($+$3 to $+$5pp).
In contrast, Qwen3-8B's strong items barely change ($-$0.2pp) but weak and control items drop sharply ($-$8 to $-$10pp); the model attempts shortcuts indiscriminately but lacks the ability to execute them on harder variants.
Llama-3.1-8B degrades across all variants, confirming that NS prompting is harmful for models without sufficient underlying capability.
Full per-scale results are in Table~\ref{tab:full_results} (Appendix~\ref{sec:appendix_full_results}).

\begin{figure*}[t]
    \centering
    \includegraphics[width=0.55\textwidth]{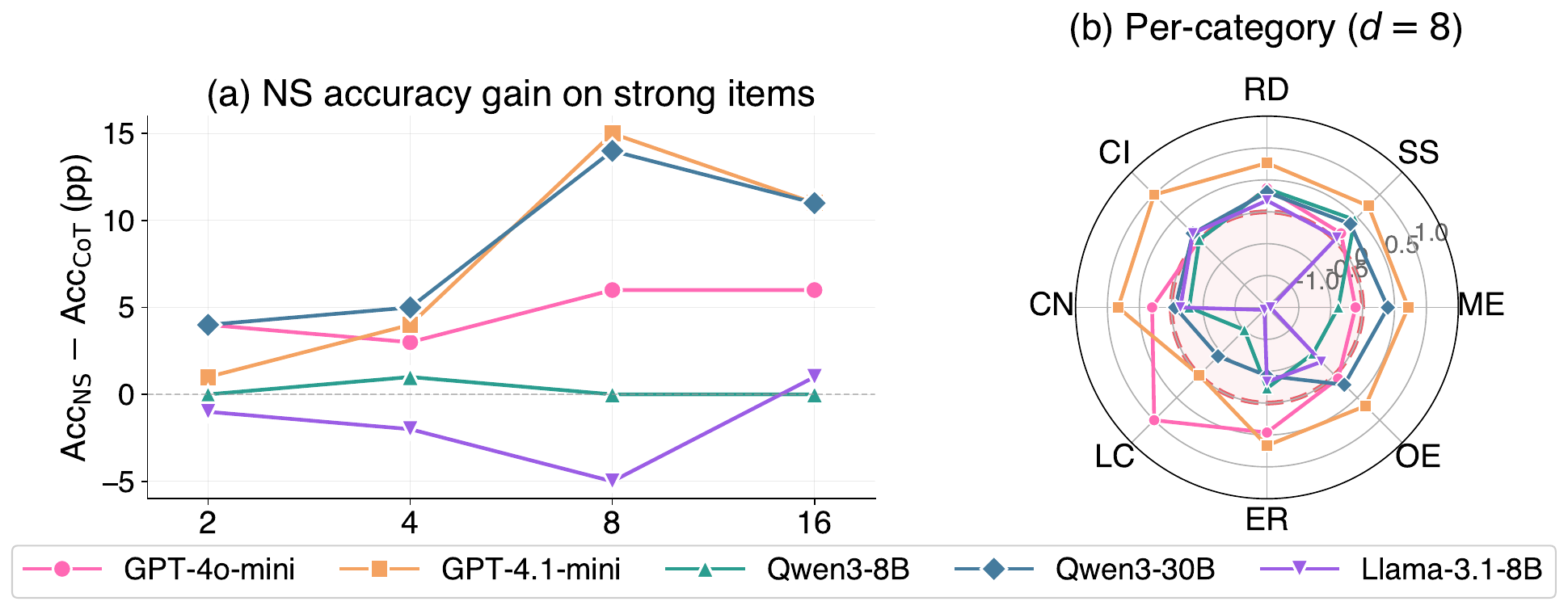}
    \caption{RQ2 analysis. (a)~NS accuracy gain on strong items increases with digit scale for capable models; 8B models show no benefit. (b)~Normalized NS improvement per category at $d{=}8$: $(\text{NS}-\text{CoT})/(1-\text{CoT})$; red dashed circle = zero baseline.}
    \label{fig:rq1_combined}
\end{figure*}

\vspace{-0.05in}
\paragraph{Per-shortcut category analysis.}
The aggregate pattern masks substantial variation across shortcut types.
Figure~\ref{fig:rq1_combined}(a) shows the normalized NS improvement $(\text{NS}-\text{CoT})/(1-\text{CoT})$ at $d{=}8$.
The largest gains come from \emph{relative distance} (RD): GPT-4.1-mini improves from 32\% to 84\% ($+$52pp), and GPT-4o-mini from 46\% to 66\% ($+$20pp)—both cases where NS prompting encourages benchmark comparison (e.g., ``both fractions are near $\nicefrac{1}{2}$'') instead of costly cross-multiplication.
\emph{Option elimination} (OE) is a revealing case: GPT-4.1-mini gains $+$26pp, yet a keyword analysis of OE responses at $d{=}4$ shows that under CoT, virtually no model uses actual elimination strategies, 
while GPT-4o-mini uses them in 0\% of CoT responses, rising to 60\% under NS; Llama-3.1-8B remains at 4\% even under NS.
This confirms that option elimination is a meta-reasoning skill that most models lack entirely and that NS prompting can partially unlock only in capable models.
By contrast, \emph{magnitude estimation} (ME) shows slight degradation for GPT-4o-mini ($-$2pp at $d{=}8$): when operands are already near round numbers, NS prompting adds no value and can introduce rounding errors.
Weaker models (Qwen3-8B, Llama-3.1-8B) show negative improvements on most categories, confirming that NS prompting hurts models that lack the capacity to execute shortcuts reliably.
Radar charts for other digit scales are provided in Appendix~\ref{sec:appendix_radar}.

\vspace{-0.1in}
\subsection{RQ3: Do LLMs Misuse Shortcuts?}
\label{sec:rq3}
\vspace{-0.1in}

\input{figures/TABLE_judge.tex}

\textbf{Models often overapply shortcuts.} As shown in Table~\ref{tab:judge}, 
most models show strong YES bias in J1: GPT-4o-mini achieves 100\% on strong items but 0\% on controls; it says YES to every item. Qwen3-8B (98\% strong, 0\% control) shows the same pattern. This reflects \emph{over-rationalisation}: models can always construct a plausible shortcut narrative, making them unable to reject control items.
In J2, where models classify whether a \emph{given solution} used a shortcut, performance is much higher (91--100\% for most models).
This asymmetry reveals that models can \emph{identify} a shortcut after the fact but cannot reliably \emph{predict} when one is appropriate.
(Evaluation setting of judgment is presented in   Appendix~\ref{sec:appendix_su_prompt}, and 
~\ref{sec:appendix_human_validation}).
\vspace{-0.1in}
\subsection{RQ4: Can LLMs Generate Shortcut-Amenable Problems?}
\label{sec:rq4}\vspace{-0.1in}


\input{figures/TABLE_generate.tex}

\begin{figure}[t]
    \centering
    \includegraphics[width=0.5\columnwidth]{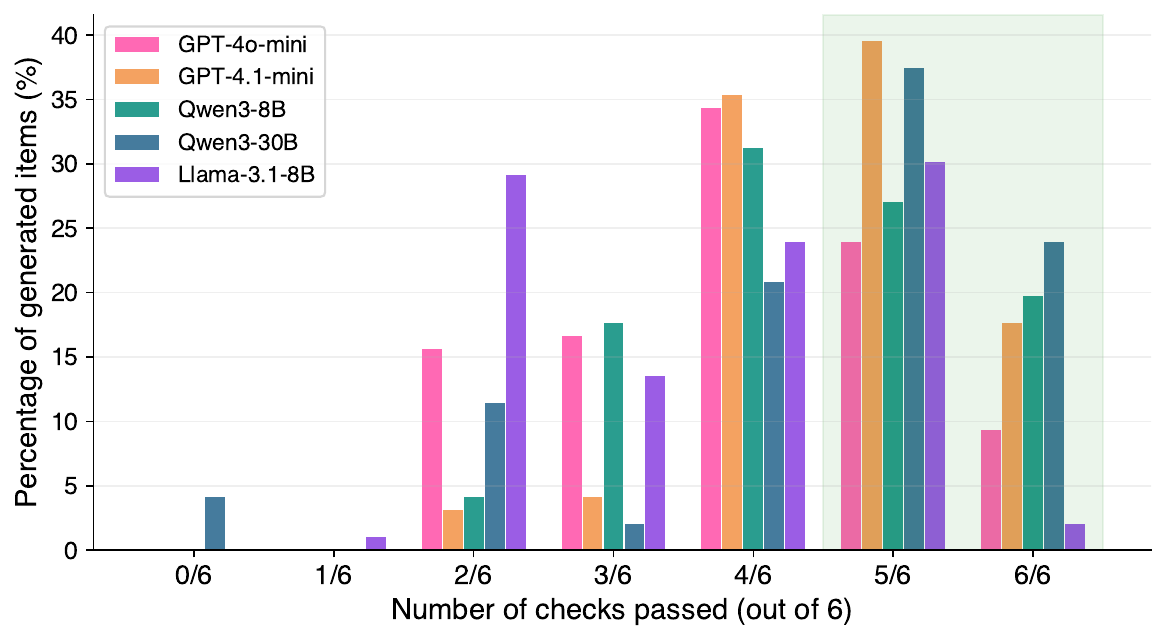} \vspace{-0.15in}
    \caption{Distribution of checks passed per generated item. Most items pass 4--5 of 6 checks, suggesting models are close to generating valid problems but consistently fail on the shortcut-existence constraint.}
    \label{fig:g_pass_dist}
\end{figure}

\textbf{Generation largely fails.}
Table~\ref{tab:generate} shows that overall pass-all-6-checks rates are low (2--24\%), but Figure~\ref{fig:g_pass_dist} reveals that the low pass rates are partly an artefact of requiring \emph{all} six checks to pass simultaneously.
A large fraction of generated items pass 4--5 of 6 checks: 58\% for GPT-4o-mini, 75\% for GPT-4.1-mini, 58\% for Qwen3-8B, 58\% for Qwen3-30B, and 54\% for Llama-3.1-8B.
The dominant bottleneck is the \emph{shortcut-existence} check (SC.Ex), which fails on 33--62\% of items across models (far more than others).
Models generate operands that \emph{look} round (e.g., $4{,}800 \times 2{,}100$) but do not simplify to trivially executable mental arithmetic, indicating that models grasp the surface form of number-sense problems but not the structural constraint that makes a shortcut effective.
The second most common failure is \emph{answer correctness} (S.Ans/C.Ans: 30--54\%), where models confuse their own shortcut estimates with exact answers.
Appendix~\ref{sec:appendix_g_examples} presents side-by-side raw outputs with check-level annotations.

\paragraph{Implications from Evaluation.}
The Apply $\gg$ Analyze $>$ Create ordering suggests that current LLMs' number sense is largely \emph{procedural}: they can execute shortcuts when instructed, but lack the declarative understanding needed to judge applicability or construct new instances.
This dissociation parallels findings in human cognition where procedural fluency develops ahead of conceptual understanding \citep{rittle2001developing}.

\vspace{-0.1in}
\subsection{RQ5: What Benefits Could NS Post-Training Bring?}
\label{sec:rq5}\vspace{-0.1in}

We evaluate Qwen3-8B on MATH-500 \citep{hendrycks2021measuring}, splitting problems into NS-amenable (171 problems, 34\%) and computation-required (329 problems, 66\%) subsets via GPT-4.1-mini classification (prompt in Appendix~\ref{sec:appendix_math500_prompt}).
On the NS-amenable subset, the base model under NS prompting already matches CoT accuracy (65.0\% vs.\ 64.5\%) while using 10\% fewer tokens (340 vs.\ 376).
\vspace{-0.1in}

\paragraph{NS Post-training is helpful.}
We fine-tune Qwen3-8B and Llama-3.1-8B with DPO on 500 MATH problems, pairing NS-style shortcut solutions as preferred responses against verbose CoT solutions.
For Qwen3-8B, the best variant (Combined-DPO) achieves 68.0\% accuracy under NS prompting ($+$3.0pp over base) without degrading generalisation across 7 OOD benchmarks (all within $\pm$4pp; Tables~\ref{tab:training} and~\ref{tab:ood} in Appendix~\ref{sec:appendix_training}).
For Llama-3.1-8B, Combined-SFT yields a larger gain ($+$16pp under CoT), but from a much lower baseline (21.5\%$\to$37.5\%), suggesting that weaker models benefit more from explicit training than from prompting alone.

%% file: figures/TABLE_judge.tex
\begin{table}[t]
\centering
\scriptsize
\caption{Applicability judgment task results. J1: shortcut appropriateness (251 items after filtering). J2: strategy identification (80 items, GPT models only).} \vspace{-0.1in}
\label{tab:judge}
\begin{tabular}{llrrr}
\toprule
Task & Model & Overall (\%) & Strong (\%) & Control (\%) \\
\midrule
J1 & GPT-4o-mini  & 66 & 100 & 0 \\
   & GPT-4.1-mini & 65 & 89 & 16 \\
   & Qwen3-8B     & 65 & 98 & 0 \\
   & Qwen3-30B    & 64 & 97 & 4 \\
   & Llama-3.1-8B & 61 & 83 & 21 \\
\midrule
 & & Overall (\%) & SHORTCUT (\%) & COMP (\%) \\
\midrule
J2 & GPT-4.1-mini & 91 & 91 & 92 \\
   & GPT-4o-mini  & 70 & 94 & 58 \\
\bottomrule
\end{tabular}\vspace{-0.1in}
\end{table}

%% file: figures/TABLE_generate.tex
\begin{table}[t]
\centering
\scriptsize
\caption{Problem generation: per-check pass rates (\%) and overall pass-all-6-checks rate. All models evaluated on 96 prompts. Fmt = valid JSON format; S.Ans = strong answer correct; C.Ans = control answer correct; SC.Ex = shortcut exists in strong; C.Blk = control blocks shortcut; Var = variant matching (same template/scale).} \vspace{-0.1in}
\label{tab:generate}
\setlength{\tabcolsep}{3pt}
\begin{tabular}{l rrrrrr r}
\toprule
Model & Fmt & S.Ans & C.Ans & SC.Ex & C.Blk & Var & Pass \\
\midrule
Qwen3-30B    & 96 & 65 & 73 & 66 & 87 & 76 & 23/96 \\
Qwen3-8B     & 100 & 69 & 64 & 55 & 83 & 71 & 19/96 \\
GPT-4.1-mini & 100 & 70 & 79 & 67 & 80 & 75 & 17/96 \\
GPT-4o-mini  & 100 & 58 & 48 & 46 & 82 & 73 & 9/96 \\
Llama-3.1-8B & 100 & 51 & 46 & 38 & 83 & 52 & 2/96 \\
\bottomrule
\end{tabular}
\end{table}

%% file: sections/5_conclusion.tex
\vspace{-0.05in}
\section{Conclusion}
\label{sec:conclusion}\vspace{-0.1in}
We introduced number sense as a new lens for evaluating numerical reasoning in LLMs. 
We presented \textsc{SenseMath}, a controlled benchmark of 4{,}800 items designed to isolate structure-sensitive behavior by varying digit scale and shortcut availability under a common surface format.
Our results show a clear gap between procedural and structural competence: LLMs can often execute shortcuts when explicitly prompted, but struggle to judge shortcut applicability and to generate new shortcut-amenable problems. This suggests that apparent success on arithmetic tasks may overestimate models' deeper numerical understanding. Although number-sense post-training yields improvements, substantial limitations remain.



%% file: figures/TABLE_examples.tex
\begin{table*}[h]
\centering
\small
\caption{Example \textsc{SenseMath} items ($d{=}2$). Each row shows one item with its strong and control variants; the strong variant admits a clean shortcut while the control requires direct computation. Correct answers are \textbf{bolded}.}
\label{tab:examples}
\begin{tabular}{@{}p{1.5cm}p{5.7cm}p{5.7cm}@{}}
\toprule
\textbf{Category} & \textbf{Strong-shortcut variant} & \textbf{Control variant} \\
\midrule
Structural (SS) &
\emph{What is $98 \times 34$?} \newline
(A)~3322 \quad (B)~3342 \quad \textbf{(C)~3332} \quad (D)~3352 \newline
{\footnotesize Shortcut: $98{=}(100{-}2)$, so $100{\times}34 - 2{\times}34 = 3332$.} &
\emph{What is $47 \times 40$?} \newline
(A)~1870 \quad (B)~1890 \quad (C)~1900 \quad \textbf{(D)~1880} \newline
{\footnotesize No near-100 structure; direct multiplication required.} \\
\addlinespace
Cancellation (CI) &
\emph{Evaluate: $71 + 28 - 27$.} \newline
(A)~82 \quad (B)~92 \quad \textbf{(C)~72} \quad (D)~62 \newline
{\footnotesize Shortcut: $28{-}27{=}1$, so $\approx 71{+}1 = 72$.} &
\emph{Evaluate: $71 + 28 - 118$.} \newline
(A)~$-19$ \quad (B)~$-1$ \quad \textbf{(C)~$-11$} \quad (D)~$-31$ \newline
{\footnotesize No near-cancellation; full arithmetic needed.} \\
\addlinespace
Equation (ER) &
\emph{$23 + 22 = \rule{0.5cm}{0.4pt} + 18$.} \newline
(A)~37 \quad \textbf{(B)~27} \quad (C)~47 \quad (D)~17 \newline
{\footnotesize Shortcut: move 18 left, $23{+}22{-}18 = 27$.} &
\emph{$20 + 93 + \rule{0.5cm}{0.4pt} = 59 + 49$.} \newline
(A)~15 \quad (B)~$-15$ \quad \textbf{(C)~$-5$} \quad (D)~5 \newline
{\footnotesize No direct identity; must compute both sides.} \\
\addlinespace
Option Elim (OE) &
\emph{What is $70 \times 16$?} \newline
(A)~1123 \quad (B)~1119 \quad (C)~1121 \quad \textbf{(D)~1120} \newline
{\footnotesize Shortcut: $70{\times}\text{even} \Rightarrow$ ends in 0; only 1120 qualifies.} &
\emph{What is $33 \times 87$?} \newline
(A)~3031 \quad (B)~2973 \quad (C)~3187 \quad \textbf{(D)~2871} \newline
{\footnotesize All options plausible; no elimination possible.} \\
\bottomrule
\end{tabular}
\end{table*}

%% file: figures/TABLE_full_results.tex
\begin{table*}[h]
  \centering
  \small
  \setlength{\tabcolsep}{2.2pt}
  \caption{Full results across all prompting conditions and digit scales. Acc$\uparrow$ (\%) = accuracy; SU = shortcut usage rate (proportion) classified by GPT-4.1-mini judge. $S$ = strong-shortcut, $W$ = weak-shortcut, $C$ = control. {\color{green!50!black}Green}/{\color{red!70!black}red} deltas on Acc\textsubscript{NS} show the change from Acc\textsubscript{CoT} on strong items. Open-weight models truncate under Strict at 512 tokens.}
  \label{tab:full_results}
  \begin{tabular}{ll l|rrr|rrr|rrr|rrr}
    \toprule
    \textbf{Model} & \textbf{Prompt} & \textbf{Metric} & \multicolumn{3}{c|}{$d{=}2$} & \multicolumn{3}{c|}{$d{=}4$} & \multicolumn{3}{c|}{$d{=}8$} & \multicolumn{3}{c}{$d{=}16$} \\
     & & & $S$ & $W$ & $C$ & $S$ & $W$ & $C$ & $S$ & $W$ & $C$ & $S$ & $W$ & $C$ \\
    \midrule
    GPT-4o-mini
     & CoT & Acc$\uparrow$ (\%) & 88 & 88 & 85 & 78 & 70 & 65 & 75 & 62 & 55 & 67 & 53 & 46 \\
     & & SU & .28 & .18 & .24 & .27 & .16 & .18 & .36 & .21 & .22 & .37 & .27 & .22 \\
    \rowcolor{gray!10}
     & NS & Acc$\uparrow$ (\%) & 92{\scriptsize\color{green!50!black}\,+4} & 92 & 87 & 84{\scriptsize\color{green!50!black}\,+6} & 72 & 63 & 82{\scriptsize\color{green!50!black}\,+7} & 73 & 62 & 70{\scriptsize\color{green!50!black}\,+3} & 59 & 51 \\
    \rowcolor{gray!10}
     & & SU & .46 & .37 & .44 & .77 & .65 & .78 & .84 & .76 & .80 & .90 & .80 & .90 \\
     & Strict & Acc$\uparrow$ (\%) & 94 & 96 & 92 & 83 & 78 & 83 & 70 & 58 & 48 & 68 & 58 & 45 \\
     & & SU & .02 & .02 & .01 & .03 & .00 & .00 & .08 & .00 & .00 & .12 & .02 & .00 \\
    \midrule
    GPT-4.1-mini
     & CoT & Acc$\uparrow$ (\%) & 99 & 99 & 99 & 88 & 89 & 76 & 80 & 72 & 65 & 74 & 58 & 60 \\
     & & SU & .16 & .07 & .07 & .20 & .04 & .01 & .46 & .22 & .37 & .57 & .34 & .48 \\
    \rowcolor{gray!10}
     & NS & Acc$\uparrow$ (\%) & 100{\scriptsize\color{green!50!black}\,+1} & 100 & 100 & 96{\scriptsize\color{green!50!black}\,+8} & 94 & 83 & 95{\scriptsize\color{green!50!black}\,+15} & 86 & 83 & 86{\scriptsize\color{green!50!black}\,+12} & 70 & 69 \\
    \rowcolor{gray!10}
     & & SU & .70 & .58 & .62 & .68 & .50 & .70 & .80 & .66 & .78 & .87 & .86 & .87 \\
     & Strict & Acc$\uparrow$ (\%) & 100 & 99 & 100 & 91 & 90 & 90 & 60 & 54 & 29 & 55 & 35 & 36 \\
     & & SU & .06 & .00 & .00 & .09 & .00 & .00 & .18 & .01 & .00 & .26 & .13 & .26 \\
    \midrule
    Qwen3-8B
     & CoT & Acc$\uparrow$ (\%) & 97 & 99 & 98 & 74 & 70 & 65 & 70 & 62 & 59 & 56 & 48 & 50 \\
     & & SU & .30 & .17 & .14 & .37 & .16 & .25 & .55 & .26 & .44 & .56 & .36 & .46 \\
    \rowcolor{gray!10}
     & NS & Acc$\uparrow$ (\%) & 98{\scriptsize\color{green!50!black}\,+1} & 98 & 95 & 72{\scriptsize\color{red!70!black}\,$-$2} & 55 & 53 & 70{\scriptsize\color{gray}\,$\pm$0} & 55 & 52 & 56{\scriptsize\color{gray}\,$\pm$0} & 38 & 34 \\
    \rowcolor{gray!10}
     & & SU & .66 & .56 & .55 & .86 & .72 & .88 & .87 & .79 & .90 & .90 & .88 & .98 \\
     & Strict & Acc$\uparrow$ (\%) & 91 & 92 & 92 & 53 & 46 & 44 & 37 & 34 & 41 & 33 & 28 & 26 \\
     & & SU & .01 & .00 & .00 & .06 & .00 & .01 & .08 & .00 & .00 & .20 & .04 & .06 \\
    \midrule
    Qwen3-30B
     & CoT & Acc$\uparrow$ (\%) & 100 & 100 & 98 & 68 & 58 & 56 & 57 & 44 & 47 & 48 & 30 & 31 \\
     & & SU & .26 & .15 & .15 & .39 & .22 & .33 & .60 & .39 & .47 & .59 & .42 & .52 \\
    \rowcolor{gray!10}
     & NS & Acc$\uparrow$ (\%) & 100{\scriptsize\color{gray}\,$\pm$0} & 100 & 100 & 73{\scriptsize\color{green!50!black}\,+5} & 68 & 56 & 74{\scriptsize\color{green!50!black}\,+17} & 50 & 52 & 60{\scriptsize\color{green!50!black}\,+12} & 42 & 44 \\
    \rowcolor{gray!10}
     & & SU & .86 & .77 & .76 & .86 & .84 & .92 & .89 & .84 & .83 & .94 & .96 & .98 \\
     & Strict & Acc$\uparrow$ (\%) & 89 & 89 & 92 & 26 & 25 & 20 & 24 & 24 & 24 & 34 & 28 & 34 \\
     & & SU & .00 & .00 & .00 & .02 & .00 & .00 & .11 & .04 & .00 & .20 & .10 & .05 \\
    \midrule
    Llama-3.1-8B
     & CoT & Acc$\uparrow$ (\%) & 82 & 76 & 80 & 66 & 70 & 58 & 62 & 53 & 51 & 51 & 36 & 33 \\
     & & SU & .26 & .24 & .25 & .31 & .25 & .24 & .37 & .24 & .24 & .40 & .33 & .26 \\
    \rowcolor{gray!10}
     & NS & Acc$\uparrow$ (\%) & 74{\scriptsize\color{red!70!black}\,$-$8} & 70 & 68 & 59{\scriptsize\color{red!70!black}\,$-$7} & 56 & 55 & 57{\scriptsize\color{red!70!black}\,$-$5} & 48 & 50 & 55{\scriptsize\color{green!50!black}\,+4} & 45 & 42 \\
    \rowcolor{gray!10}
     & & SU & .62 & .58 & .59 & .78 & .76 & .69 & .86 & .82 & .90 & .92 & .87 & .86 \\
     & Strict & Acc$\uparrow$ (\%) & 55 & 62 & 61 & 46 & 50 & 40 & 36 & 36 & 30 & 22 & 23 & 20 \\
     & & SU & .00 & .00 & .00 & .00 & .00 & .00 & .01 & .00 & .00 & .04 & .00 & .00 \\
    \bottomrule
  \end{tabular}
\end{table*}

%% file: figures/TABLE_training.tex
\begin{table}[t]
\centering
\caption{In-domain results on the 200-problem eval set (500-example training). SU Rate = fraction of responses using shortcut strategy; Avg Tok = average response tokens. Best accuracy per model per prompt condition is \textbf{bold}.}
\label{tab:training}
\small
\begin{tabular}{ll l ccc}
\toprule
Model & Variant & Prompt & SU Rate & Accuracy & Avg Tok \\
\midrule
Qwen3-8B   & Base          & CoT & 10.0\% & 64.5\%          & 376 \\
(nothink)  & Base          & NC  & 36.5\% & 65.0\%          & 340 \\
           & NS-DPO        & CoT & 11.5\% & \textbf{67.0\%} & 372 \\
           & NS-DPO        & NC  & 33.5\% & 64.0\%          & 350 \\
           & Baseline-DPO  & CoT &  8.5\% & 66.0\%          & 374 \\
           & Baseline-DPO  & NC  & 34.5\% & 60.0\%          & 350 \\
           & NS-SFT        & CoT & 11.5\% & 66.0\%          & 369 \\
           & NS-SFT        & NC  & 32.5\% & 64.0\%          & 348 \\
           & CoT-SFT       & CoT & 11.5\% & 64.0\%          & 373 \\
           & CoT-SFT       & NC  & 34.5\% & 63.0\%          & 351 \\
           & Combined-SFT  & CoT & 11.5\% & 63.0\%          & 392 \\
           & Combined-SFT  & NC  & 22.0\% & 65.5\%          & 372 \\
           & Combined-DPO  & CoT & 11.0\% & 62.5\%          & 395 \\
           & Combined-DPO  & NC  & 23.5\% & \textbf{68.0\%} & 375 \\
\midrule
Llama-3-8B & Base          & CoT &  5.5\% & 21.5\%          & 363 \\
           & Base          & NC  & 44.5\% & 15.5\%          & 269 \\
           & NS-DPO        & CoT &  3.5\% & 22.5\%          & 361 \\
           & NS-DPO        & NC  & 48.0\% & 17.0\%          & 271 \\
           & Baseline-DPO  & CoT &  4.5\% & 18.5\%          & 365 \\
           & Baseline-DPO  & NC  & 51.0\% & 17.0\%          & 268 \\
           & NS-SFT        & CoT &  3.5\% & 24.0\%          & 398 \\
           & NS-SFT        & NC  & 57.5\% & 16.0\%          & 285 \\
           & CoT-SFT       & CoT &  3.0\% & 25.0\%          & 379 \\
           & CoT-SFT       & NC  & 59.0\% & 19.0\%          & 292 \\
           & Combined-SFT  & CoT &  4.0\% & \textbf{37.5\%} & 321 \\
           & Combined-SFT  & NC  & 16.0\% & 31.5\%          & 286 \\
           & Combined-DPO  & CoT &  4.5\% & 36.5\%          & 325 \\
           & Combined-DPO  & NC  & 12.0\% & 30.0\%          & 288 \\
\bottomrule
\end{tabular}
\end{table}

%% file: figures/TABLE_ood.tex
\begin{table}[t]
\centering
\caption{Out-of-distribution benchmark results after 500-example training. Most variants fall within $\pm$4pp of base, confirming no catastrophic forgetting.}
\label{tab:ood}
\small
\setlength{\tabcolsep}{3.5pt}
\begin{tabular}{ll ccccccc}
\toprule
Model & Variant & BBH & ARC & GPQA & MMLU-P & MedQA & LogiQA & CSQA \\
\midrule
Qwen3-8B & Base         & 73.8 & 87.0 & 34.6 & 31.8 & 61.4 & 50.5 & 80.8 \\
         & NS-DPO       & 73.8 & 87.0 & 31.2 & 31.4 & 59.4 & 46.9 & 80.8 \\
         & Baseline-DPO & 73.6 & 86.6 & 34.6 & 31.8 & 59.0 & 47.9 & 80.7 \\
         & NS-SFT       & 73.6 & 86.6 & 33.7 & 32.6 & 60.4 & 48.4 & 80.7 \\
         & CoT-SFT      & 73.4 & 86.3 & 32.8 & 30.6 & 60.6 & 48.7 & 80.9 \\
         & Comb-DPO     & 69.8 & 87.0 & 35.0 & 28.4 & 62.2 & 51.0 & 80.0 \\
\midrule
Llama-3-8B & Base       & 50.2 & 80.3 & 29.2 & 19.2 & 60.6 & 41.6 & 75.4 \\
           & NS-DPO     & 49.8 & 80.6 & 28.6 & 19.4 & 59.6 & 42.4 & 75.7 \\
           & Baseline-DPO & 50.2 & 80.3 & 28.8 & 18.4 & 60.2 & 42.1 & 75.7 \\
           & NS-SFT     & 50.6 & 80.3 & 28.8 & 18.8 & 60.4 & 42.4 & 75.8 \\
           & CoT-SFT    & 50.4 & 80.3 & 29.9 & 19.0 & 60.4 & 41.8 & 75.9 \\
           & Comb-SFT   & 54.4 & 82.6 & 27.9 & 17.4 & 59.4 & 40.9 & 74.6 \\
           & Comb-DPO   & 54.4 & 82.3 & 29.2 & 18.0 & 59.6 & 41.0 & 74.5 \\
\bottomrule
\end{tabular}
\end{table}